\pdfoutput=1
\documentclass[11pt,a4paper]{article}
\usepackage[hyperref]{acl2021}
\usepackage{times}
\usepackage{latexsym}

\usepackage{microtype}

\usepackage{algorithm}
\usepackage[noend]{algpseudocode}
\usepackage{amsmath}
\usepackage{amssymb}
\usepackage{booktabs}
\usepackage{mathtools}
\usepackage{pgfplots}
\usepackage{stmaryrd}
\usepackage{subfig}
\usepackage{tabularx}
\usepackage{tikz}

\DeclareMathOperator{\softmax}{softmax}
\DeclareMathOperator{\relu}{ReLU}
\DeclarePairedDelimiter\abs{\lvert}{\rvert}

\usetikzlibrary{arrows.meta}
\tikzset{>=latex}

\pgfplotsset{compat=newest}
\pgfplotsset{every axis/.append style={xticklabel style={font=\small}
, yticklabel style={font=\small}}}
\usepgfplotslibrary{colorbrewer}

\aclfinalcopy
\def\aclpaperid{} %  Enter the acl Paper ID here

\newcommand{\rf}{\textsc{Ref}}
\newcommand{\cai}{\textsc{CL20}}
\newcommand{\chrf}{\textsc{chrF++}}
\newcommand{\tdms}{\textsc{Tr-D}}
\newcommand{\tdsd}{\textsc{Tr-M}}
\newcommand{\tdmd}{\textsc{Tr-S}}
\newcommand{\tdmsd}{\textsc{Tr}}
\newcommand{\ldca}{LDC2017T10}
\newcommand{\ldcb}{LDC2015E86}
\newcommand{\rowgr}[1]{\hspace{-1em}#1}

\title{Tree Decomposition Attention for AMR-to-Text Generation}

\author{Lisa Jin \and Daniel Gildea\\
  Department of Computer Science\\
  University of Rochester\\
Rochester, NY 14627}

\begin{document}
\maketitle
\begin{abstract}
Text generation from AMR requires mapping a semantic graph to a string that it annotates. Transformer-based graph encoders, however, poorly capture vertex dependencies that may benefit sequence prediction. To impose order on an encoder, we locally constrain vertex self-attention using a graph's tree decomposition. Instead of forming a full query-key bipartite graph, we restrict attention to vertices in parent, subtree, and same-depth bags of a vertex. This hierarchical context lends both sparsity and structure to vertex state updates. We apply dynamic programming to derive a forest of tree decompositions, choosing the most structurally similar tree to the AMR\@. Our system outperforms a self-attentive baseline by 1.6 BLEU and 1.8 \chrf{}.
\end{abstract}

\section{Introduction}
Text generation from structures such as graphs and tables applies widely to end user applications. For example, a goal of the SQL-to-text task may be to interpret database queries for non-experts. Such tasks often transform an unordered entity set (e.g., semantic graph vertices) into a token sequence. To achieve this, recent models often employ global self-attention over input elements. However, the lack of local structural constraints may hurt a model's ability to transduce between input and output structure---a major concern for complex inputs such as graphs.

A generated string both simplifies and expands upon a given semantic graph. On the one hand, a string reduces the graph topology into a chain. On the other hand, the string's tokens can specify syntactic roles of semantic graph vertices. These dual properties suggest that a model must impose order on the graph while providing lexical details.

\begin{figure}
  \centering
    \begin{tikzpicture}
      \begin{scope}[font=\footnotesize\sffamily,
        every node/.style={circle,thick,draw,inner sep=0.125em}]
        \node (a) at (0,0) [label={[inner sep=0,shift={(0,-.4)}]abide-01}] {};
        \node (w) at (-1,-1) [label={[inner sep=0,shift={(-.6,-.4)}]want-01}] {};
        \node (t) at (1,-1) [label={[inner sep=0,shift={(.4,-.3)}]they}] {};
        \node (p) at (-2,-2) [label={[inner sep=0,shift={(-.6,-.4)}]post-01}] {};
        \node (t2) at (-3,-3) [label={[inner sep=0,shift={(-.5,-.3)}]there}] {};

        \node (y) at (0,-3) [label={[inner sep=0,shift={(.5,-.3)}]you}] {};
      \end{scope}
      \begin{scope}[font=\footnotesize\ttfamily,>=latex,auto,
        every node/.style={fill=none,circle,inner sep=0},
        every edge/.style={draw=black,thin}]
        \path [->] (a) edge node [shift={(.2,0)}] {arg1} (t);
        \path [->] (a) edge node [shift={(-1.5,.9)}] {condition} (w);
        \path [->,bend left=10] (a) edge node [shift={(.2,-.2)}] {arg0} (y);
        \path [->] (w) edge node [shift={(-.7,.5)}] {arg1} (p);
        \path [->,bend right=10] (w) edge node [shift={(0,.3)}] {arg0} (y);
        \path [->] (p) edge node [shift={(-.7,.5)}] {arg2} (t2);
        \path [->,bend right=10] (p) edge node [shift={(-.4,.2)}] {arg0} (y);
      \end{scope}
    \end{tikzpicture}
  \caption{AMR for the sentence, ``If you want to post there you abide by them.''}
  \label{fig:sample-amr}
\end{figure}
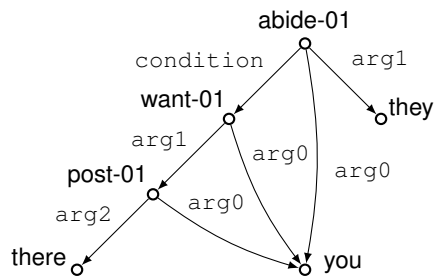

Abstract Meaning Representation \cite{banarescu2013abstract} is a popular semantic formalism that produces rooted, directed graphs. Its labeled vertices and edges, called concepts and relations, are largely based on PropBank framesets \cite{kingsbury2002treebank}. Each concept belongs to a known frame; e.g., `observe-01' has a slot for the observer (:arg0) and one for the observed entity (:arg1). Figure \ref{fig:sample-amr} is a sample AMR with a reentrancy, or vertex with multiple parents, at the concept `you'.

Among neural AMR-to-text models, graph encoders based on the Transformer \cite{vaswani2017attention} have gained acceptance over graph neural networks (GNNs). In each GNN iteration, vertex state is updated as a function of messages from neighboring vertices. Transformer-style encoders remove this locality constraint by allowing self-attention over all graph vertices. Whereas GNN-based models \citep{song2018graph,beck2018graph} often apply RNNs along successive vertex updates, Transformer-style ones \citep{zhu2019modeling,cai2020graph} stack attention layers. The improved performance of the latter models may be due to their clique-like view of vertex communication.

The expressivity of self-attention may come at the price of less targeted vertex updates. Without the restrictions of neighborhood-level updates, vertex states may lack local graph context helpful for sequence prediction. Other forms of AMR-to-text models \cite{wang2020amr,yao2020heterogeneous} are based on graph attention networks \cite{velivckovic2017graph}. They adopt GNN-style filters over vertex neighborhoods to constrain self-attention. The better performing system by \citeauthor{yao2020heterogeneous}\ encodes views of the input graph at varying connectivities (e.g., bidirectional edges, reverse edges only, fully connected). We explore whether a model can integrate these types of vertex dependencies without curated graph views.

Our main approach is to bias neural graph encoders on graph-theoretic structures called tree decompositions. Each tree node covers an edge-disjoint subgraph of the graph. Tree decompositions also correspond to factor graphs (Figure~\ref{fig:td}), where nodes map to factors. As such, they offer a compromise between local and global levels of graph connectivity. Past work by \citet{jones2013modeling} describes how rules of a context-free graph grammar can be extracted from a given tree decomposition. Despite the ostensible utility of such grammars, the authors do not apply them to tasks involving semantic graphs.

This work (i) automatically extracts structure-aligned tree decompositions from AMRs and (ii) conditions a graph encoder on them for text generation. To achieve (i) we adapt bottom-up graph grammar parsing to tree decompositions. For (ii) we inject the topology and intra-node context of these trees into vertex-level attention. We hypothesize that the structure and sparsity of tree decomposition attention will improve model generalization.

\section{Tree Decomposition}
Here we define tree decomposition and the notion of treewidth. We then show how tree decomposition relates to text generation through semantic parsing. Lastly, we describe an $\mathcal{O}(n^{k + 1})$ algorithm to find all $k$-width tree decompositions of a graph.

\subsection{Definition}
\label{sec:td-def}
A tree decomposition partitions a graph $G = (V, E)$ into a tree $T = (I, F)$ of overlapping vertex sets $\{X_i\}_{i \in I}$ called bags. Bags are connected via arcs $F$ such that all bags $X_k$ along the path between $X_i$ and $X_j$ contain the vertices $X_i \cap X_j$---a property called \textit{running intersection}. Other properties include \textit{vertex cover} $\bigcup_i X_i = V$ and \textit{edge cover}, where each $(u, v) \in E$ resides in exactly one bag.

Graphs that are more tightly interconnected generally require
larger bags. The width of a tree decomposition (henceforth TD),
$\max_i |X_i| - 1$, depends on its largest bag size. To describe
the complexity of $G$, we define \textit{treewidth} as the minimum width
across all of its TDs.  The treewidth of an acyclic graph is one, and
low treewidth in general indicates that the graph has a tree-like
structure.

Summarizing a graph with a low-width TD can improve algorithmic
efficiency.  Many dynamic programming algorithms take time linear in
the size of the graph for graphs of constant treewidth, including
the widely used junction tree algorithm for probabilistic graphical models
\cite{Arnborg:1991}.

\subsection{Relation to transition-based parsing}
A parser can exploit the same bag-level structure as a tree decomposition to incrementally extract graphs from strings. A stack-based transition system can transfer tokens from a buffer to a stack, considering whether each token $i$ is linked to tokens $1 \ldots i - 1$ on the stack. While \citet{covington2001fundamental} intended for this algorithm to parse dependency trees, it easily generalizes to graphs. The parser runs in time $\mathcal{O}(n^2)$ as it processes entries of a lower triangular adjacency matrix. To improve efficiency, \citet{gildea2018cache} introduce a transition system that produces graphs of maximum treewidth $k$ in time $\mathcal{O}(n(k + 1))$. This parser replaces the stack with a cache of size $m = k + 1$. Its cache states form a TD of the output graph, where a \textit{push} action creates a new bag and \textit{pop} returns to the current bag's parent.

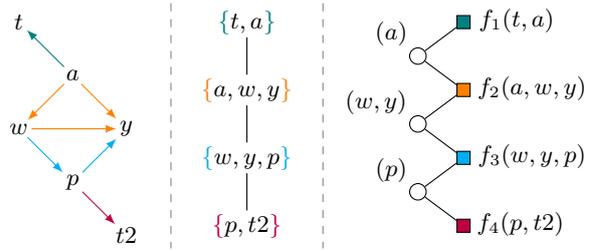
\begin{figure}
  \centering
  \begin{tikzpicture}
    \begin{scope}[every node/.style={inner sep=.1em},font=\small]
      \node (t) {$t$};
      \node (a) [below right of=t] {$a$};
      \node (w) [below left of=a] {$w$};
      \node (y) [below right of=a] {$y$};
      \node (p) [below left of=y] {$p$};
      \node (t2) [below right of=p] {$t2$};
      \node (w2) at (3.,0) {$\textcolor{teal}{\{}t,a\textcolor{teal}{\}}$};
      \node (x) at (3.,-.9) {$\textcolor{orange}{\{}a,w,y\textcolor{orange}{\}}$};
      \node (y2) at (3.,-1.8) {$\textcolor{cyan}{\{}w,y,p\textcolor{cyan}{\}}$};
      \node (z) at (3.,-2.7) {$\textcolor{purple}{\{}p,t2\textcolor{purple}{\}}$};
    \end{scope}
    \begin{scope}[>=latex]
      \path [->,color=teal] (a) edge node {} (t);
      \path [->,color=orange] (a) edge node {} (w);
      \path [->,color=orange] (a) edge node {} (y);
      \path [->,color=orange] (w) edge node {} (y);
      \path [->,color=cyan] (w) edge node {} (p);
      \path [->,color=cyan] (p) edge node {} (y);
      \path [->,color=purple] (p) edge node {} (t2);
    \end{scope}
    \begin{scope}[every node/.style={inner sep=.225em,draw},font=\small,every node/.label/.style={inner sep=0},font=\small]
      \node [fill=teal] (w3) at (5.85,0) [label={[shift={(0.7,-.325)}]$f_1(t, a)$}] {};
      \node [fill=orange] (x3) at (5.85,-.9) [label={[shift={(.9,-.325)}]$f_2(a,w,y)$}] {};
      \node [fill=cyan] (y3) at (5.85,-1.8) [label={[shift={(.9,-.325)}]$f_3(w,y,p)$}] {};
      \node (z2) [fill=purple] at (5.85,-2.7) [label={[shift={(.75,-.325)}]$f_4(p,t2)$}] {};
    \end{scope}
    \begin{scope}[every node/.style={inner sep=.2em,draw,circle},font=\small]
      \node (v0) at (5.25,-.45) [label={[shift={(-.35,-.2)}]$(a)$}] {};
      \node (v1) at (5.25,-1.35) [label={[shift={(-.55,-.375)}]$(w,y)$}] {};
      \node (v2) at (5.25,-2.25) [label={[shift={(-.35,-.2)}]$(p)$}] {};
    \end{scope}
    \draw [color=gray,style=dashed] (2,.25) -- (2,-3);
    \draw [color=gray,style=dashed] (4,.25) -- (4,-3);
    \draw (w3) -- (v0) -- (x3) -- (v1) -- (y3) -- (v2) -- (z2);
    \draw (w2) -- (x) -- (y2) -- (z);
  \end{tikzpicture}
  \caption{Simplified AMR (left), 2-width TD (center), and factor graph view of the TD (right).}
  \label{fig:td}
\end{figure}

The cache transition system exploits the low treewidth of the AMR relative to input word order. \citeauthor{gildea2018cache} compute the mean treewidths relative to optimal, string, and random order token traversals as 1.52, 2.80, and 4.84 on the LDC2015E86 corpus. Since optimal and string TDs have low treewidth, we posit that the former can help bridge between AMRs and sentences. Figure \ref{fig:td} maps from subgraphs of the AMR in Figure \ref{fig:sample-amr} to TD bags. Note that vertices $y, w, p, t2$ for the phrase ``you want to post there'' are covered by TD bags 2--4.

\subsection{Extraction from graphs}
Since we are interested in optimal TDs of AMRs, we now describe an algorithm to find all TDs of maximum width $k$ \cite{lautemann1988decomposition}. The algorithm uses dynamic programming to decompose edge-disjoint subgraphs of the original graph. Although the output structure is a derivation forest of TDs, we augment the algorithm with a scoring function to select a single TD per AMR.

The first step is to fill a dictionary $\mathcal{D}$ keyed by graph separators, each mapping to a list of corresponding connected components from $G$. A separator is a vertex set that splits $G$ into distinct connected components after its removal. An entry of $\mathcal{D}$ takes the form $S_i\colon [C_{i1} \ldots C_{ij_i}]$, where $S$ and $C$ are vertex sets of their respective subgraphs. Note that each $C$ is adjacent to a set of vertices in its $S$, written as $\mathcal{N}_S(C) \subseteq S$. Since $C$ only interacts with the rest of graph $G$ through vertices $\mathcal{N}_S(C)$, it can be decomposed separately. This independence allows an algorithm to recurse on subgraphs of $G$, caching their states along the way.

\begin{algorithm}[ht]
  \caption{Recognition of $k$-width TD}
  \label{alg:rec-decomp}
  \begin{algorithmic}[1]
  \Require $pairs \gets \{(S_i, C_{i\cdot})\}$ from graph $G$
    \Procedure{Decomp}{$S, C, k$}
    \If{$|\mathcal{N}_S(C) \cup C| \leq k + 1$} \Return \texttt{true} \EndIf
      \State $root \gets$ \texttt{false}
      \ForAll{$S^\prime \colon \mathcal{N}_S(C) \subset S^\prime \subseteq \mathcal{N}_S(C) \cup C$
      \\\hspace{5.85em}and $|S^\prime| \leq k + 1$}
      \State $ch\_root \gets$ \texttt{true}
      \ForAll{$C^\prime \colon (S^\prime, C^\prime) \in pairs$}
      \State $ch\_root \gets ch\_root\; \wedge$
      \\\hspace{9.35em}\Call{Decomp}{$S^\prime, C^\prime, k$}
      \EndFor
      \State $root \gets root \vee ch\_root$
      \EndFor
      \State \Return $root$
    \EndProcedure
  \end{algorithmic}
\end{algorithm}

\begin{figure}
  \centering
  \begin{tikzpicture}
    \begin{scope}[every node/.style={inner sep=.1em},font=\small]
      \node [inner sep=.2em,color=blue!50] (a1) {$\{t,\boldsymbol{a}\}$};
      \node (a2) [below left of=a1] {$\boldsymbol{w}$};
      \node (a3) [below right of=a1] {$\boldsymbol{y}$};
      \node [color=red!50] (a4) [below right of=a2] {$p$};
      \node [color=red!50] (a5) [below right of=a4] {$t2$};
      \node [inner sep=.2em,color=blue!50] (b1) at (3.5,0) {$\{t,a\}$};
      \node [inner sep=.2em,color=blue!50] (b2) at (3.5,-.9) {$\{a,\boldsymbol{w},\boldsymbol{y}\}$};
      \node (b3) [below of=b2,yshift=.5em] {$\boldsymbol{p}$};
      \node [color=red!50] (b4) [below right of=b3] {$t2$};
    \end{scope}
    \begin{scope}[>=latex]
      \path [->,dashed] (a1) edge node {} (a2) edge node {} (a3);
      \draw [->] (a2) -- (a3);
      \draw [->,red!50] (a2) -- (a4);
      \draw [->,red!50] (a4) -- (a5);
      \draw [->,red!50] (a4) -- (a3);
      \draw[-Implies,double distance=1.5pt] (1.6,-1.15) -- (2,-1.15);
      \draw [blue!50] (b1) -- (b2);
      \path [->,dashed] ([xshift=6pt,yshift=-8pt]b2.west) edge node {} (b3);
      \path [->,dashed] (b3) edge node {} ([xshift=-6pt,yshift=-8pt]b2.east);
      \draw [->,red!50] (b3) -- (b4);
    \end{scope}
  \end{tikzpicture}
  \caption{A recursive step of Algorithm~\ref{alg:rec-decomp}. Left: $S = \{t,a\},\ S^\prime = \{a,w,y\}$. Right: $S = \{a,w,y\},\ S^\prime = \{w,y,p\}$. Seen bags are blue and unseen edges red. Dashed edges link vertices of $S^\prime$ with those of $\mathcal{N}_S(C)$.}
  \label{fig:rec-step}
\end{figure}
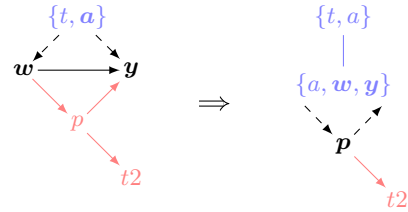

Algorithm \ref{alg:rec-decomp} recognizes whether a graph $G$ has a $k$-width TD\@. It receives as input $(S, C)$ pairs from the dictionary $\mathcal{D}$. The base case on line 2 applies when the subgraph induced by the union of $C$ and its neighbors in $S$ fits in a $k$-width TD bag. Subsequent lines recursively check whether the $(S, C)$ pair can be broken down into $(S^\prime, C^\prime)$ pairs belonging to a $k$-width TD for all $C^\prime \in \mathcal{D}[S^\prime]$. Though the algorithm can return upon \texttt{true} assignment at line 10, we keep its current form so that it may generalize to a forest of TDs. Specifically, if we view $(S, C)$ as a hyperedge, line 4 iterates over possible right-hand sides of rules in a hyperedge replacement grammar (HRG). Figure~\ref{fig:rec-step} illustrates the chained recognition of bags.

Algorithm \ref{alg:rec-decomp} is written in terms of a $(\vee, \wedge)$ semiring, which may be replaced with $(\cup, \cdot)$ to instead return a TD instance. To convert from a derivation tree of $(S, C)$ nodes to a TD, we discard the $C$ elements and merge nodes mapping to the same $S$.

\textbf{Scoring AMR TDs} In the interest of biasing self-attention on a single AMR TD, we describe a scheme to select it from among a TD forest. The general approach is to score TDs according to their adherance to the original AMR structure. AMRs are rooted graphs with directed edges pointing towards the leaves. Thus, we check whether the arcs between bags of a TD mirror the direction of edges in the AMR\@. Concretely, if TD bag $X_i$ is the parent of $X_j$ the penalty function is
\begin{align}
  \ell(X_i, X_j) &= |\{(v, u) \mid (u, v) \in E\}|,
\end{align}
where $v \in X_i - X_j$ and $u \in X_j$. In other words, edges stemming from $X_i \cap X_j$ should not have endpoints unique to parent $X_i$. This function decomposes additively across arcs of the TD due to its edge-disjoint bags. It follows that we can insert the $(\min, +)$ semiring over $\ell(\cdot, \cdot)$ into Algorithm \ref{alg:rec-decomp} to find the least penalized TD.

\section{Graph Transformer}
Our baseline \cite{cai2020graph} is a Transformer-based model designed for text generation from AMR graphs. It adapts sequential Transformers to graphs such as AMR by including relation embeddings in vertex self-attention. We describe these changes after reviewing Transformer attention.

\subsection{Encoder}
\label{sec:tform-encoder}
The encoder's aim is to learn new representations of source tokens $\mathbf{x} = (x_1, \dots, x_n)$ as a function of their initial embeddings. Denote by $X \in \mathbb{R}^{n \times d}$ the packed $d$-dimensional embeddings of $\mathbf{x}$. The encoder computes function $e\colon \mathbb{R}^{n \times d} \to \mathbb{R}^{n \times d}$ by applying dot-product attention from rows of $X$ to itself. The Transformer computes this function in parallel across all positions of $\mathbf{x}$ by operating on packed query, key, and value matrices:
\begin{align*}
  \mathrm{Attn}(X) &= \bigparallel_{h = 1}^H A^h (XV^h),\\
  A^h &= \sigma\left(\frac{1}{\sqrt{m}}(XQ^h)(XK^h)^\top\right),
\end{align*}
where $Q_h, K_h \in \mathbb{R}^{d \times m}$ and $V_h \in \mathbb{R}^{d \times d}$ are query, key, and value projections; $\bigparallel$ indicates concatenation and $\sigma$ refers to the $\softmax$ function. The $h$ superscripts index the $H$ attention heads to support a variety of projections that are later concatenated.

The $\mathrm{Attn}$ function exists within a sub-layer that also includes two linear projections and layer normalization. Multiple sub-layers are often stacked together, with residual connections in between, to allow function composition.

\textbf{Relation encoder} Given a string $\mathbf{x}$ it is possible to augment the embeddings $X$ with row-wise positional encodings. This provides the model with a notion of token ordering, which is helpful for assessing query-key compatibility. In the case of an AMR input, $\mathbf{x}$ is an unordered vertex set $V = \{v_1, \dots, v_n\}$. Due to the possiblity of cycles in the AMR, distance between vertices is ill-defined. \citeauthor{cai2020graph} respond by learning relation embeddings $\widetilde{R} \in \mathbb{R}^{n \times n \times d}$ for each entry of the adjacency matrix. A particular $\widetilde{R}_{ij}$ is the result of applying a bidirectional GRU network along the path between $v_i$ and $v_j$. The authors add reverse edges to $G$ to support all pairwise paths, as well as a global vertex $v_0$ that serves as both a source and sink in $G$. The embeddings $\widetilde{R}$ are then summed with $X$ in the attention dot-product for query $i$ and key $j$: $\big((X_i + \widetilde{R}_{ij})Q\big) \big((X_j + \widetilde{R}_{ji})K\big)^\top$. The authors claim that pairwise path embeddings offer the model a richer view of global graph structure.

\subsection{Decoder}
The decoder operates in a way analogous to the encoder by executing the function $d: \mathbb{R}^{n \times d} \to \mathbb{R}^{m \times d}$. Given the encoder output $X^\prime \in \mathbb{R}^{n \times d}$ it seeks to produce an $m$-length token sequence $\mathbf{y} = (y_1, \ldots, y_m)$. Thus, at each time step $t \in [1, m]$ it applies attention on both the decoded embeddings $Y \in \mathbb{R}^{(t - 1) \times d}$ and the encoded $X^\prime$. In the former case projections $Q, K, V$ all apply to $Y$, while in the latter $Q$ applies to $Y$ and $K, V$ apply to $X^\prime$.

To handle sparsity in the distribution of $\mathbf{y}$, \citeauthor{cai2020graph} follow previous NLG models by using a copy mechanism \cite{gu2016incorporating}. This lets the decoder mix target token generation with copied source AMR concept labels. It learns a distribution $p_{\mathrm{cp}} \in [0, 1]$ based on decoder hidden state $Y_t$:
\begin{align*}
  p(y_t | Y_t) &= ( 1 - p_{\mathrm{cp}})\hat{p}(y_t | Y_t) + p_{\mathrm{cp}} \sum_{s \in \mathcal{C}(y_t)} A_{ts},
\end{align*}
where $\hat{p}(y_t | Y_t)$ is the initial decoder target distribution, $\mathcal{C}(y_t)$ is the set of concepts corresponding to $y_t$, and $A \in m \times n$ is a cross-attention matrix between tokens and concepts.

\section{Tree Decomposition Attention}
We now discuss our modifications to encoder self-attention for text generation. The goal is use the topology of an AMR's optimal TD as a scaffold for concept-level attention.

We alter both the attention function and the relation encoder it relies on. The former imposes hierarchy onto self-attention, where vertices in bags closer to the TD root are allowed a more global range of attention. The latter attaches bag-level features to each relation embedding.

\subsection{Masked tree attention}
\label{sec:masked-tree}
In place of full pairwise vertex attention, we limit the eligible keys per query based on the TD structure. We allow each bag's vertices to attend to all vertices of its parent, descendant, and same-depth bags. Note that depth here is relative to the deepest leaf in a bag's subtree, not the root.

\begin{figure}[h]
  \centering
  \begin{tikzpicture}
    \begin{scope}[every node/.style={inner sep=.1em},font=\small]
      \node (a) {$a$};
      \node (b) [below left of=a] {$b$};
      \node (c) [below right of=a] {$c$};
      \node (d) [below left of=b] {$d$};
      \node (e) [below right of=b] {$e$};
      \node (f) [below right of=c] {$f$};
      \node (g) [below left of=e] {$g$};
    \end{scope}
    \begin{scope}[every node/.style={inner sep=.2em},font=\small]
      \node (v) at (4.5,0) {$\{a,c\}$};
      \node (w) [below left of=v] {$\{a,b\}$};
      \node (x) [below right of=v] {$\{c,f\}$};
      \node (y) [below left of=w] {$\{b,d,g\}$};
      \node (z) [below right of=w] {$\{b,e,g\}$};
    \end{scope}
    \draw (a) -- (c) -- (f);
    \draw (a) -- (b) -- (d) -- (g);
    \draw (b) -- (e) -- (g);
    \draw (b) -- (g);
    \draw (v) -- (w) -- (y);
    \draw (v) -- (x);
    \draw (w) -- (z);
  \end{tikzpicture}
  \caption{Sample graph (left) and a valid TD (right). Tree attention lets bag $\{a,b\}$ attend to all bags except $\{c,f\}$, which has a different leaf-relative depth.}
\end{figure}
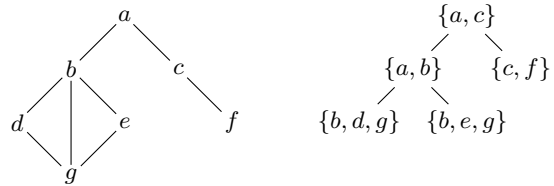

Each bag is represented by the set of concepts it contains, where attention is modeled per concept. In essence, the bag representations are distributed across AMR concepts. If a concept exists in multiple bags, it can attend to all vertices of the bag closest to the root bag. Recall that due to the running intersection property (\S\ref{sec:td-def}), the bags containing a given vertex are connected. Thus, subtree concept masks can be found by making a bottom-up pass over the AMR's TD\@. Algorithm~\ref{alg:subtree-mask} provides the steps necessary to do so.

\begin{algorithm}[h]
  \caption{Subtree mask computation}
  \label{alg:subtree-mask}
  \begin{algorithmic}[1]
    \Require TD $T = (I, F)$, concept mask $A \in \{0, 1\}^{|V| \times |V|}$
    \Procedure{SubtreeMask}{$T, A, i$}
    \ForAll{$v \in X_i$}
        \State $A_{v} \gets \bigvee_{v^\prime \in X_i} \textsc{OneHot}(v^\prime)$
        \ForAll{$X_j \in \textsc{Children}(X_i)$}
        \State $A_{v} \gets A_v \vee \bigvee_{v^\prime \in X_j} A_{v^\prime}$
        \EndFor
      \EndFor
    \EndProcedure
  \end{algorithmic}
\end{algorithm}

\subsection{Bag embeddings}
\label{sec:bag-sg}
As described in \S\ref{sec:tform-encoder}, the baseline's relation encoder uses an RNN over relation paths to learn per-relation embeddings. Here we propose to include features for the bag a relation resides in.

We make use of a bag's internal vertex-level structure and relations to form its representation. The below three features encompass a bag's structure, content, and position in the TD, respectively.

\begin{figure}[h]
  \centering
  \begin{tikzpicture}
    \begin{scope}[every node/.style={circle,thick,draw,inner sep=.125em,outer sep=0em},font=\small]
      \node (a1) at (0,0) {};
      \node (a2) at (1,0) {};
      \node (b1) at (3,.433) {};
      \node (b2) at (2.5,-.433) {};
      \node (b3) at (3.5,-.433) {};
      \node (c1) at (5.5,.433) {};
      \node (c2) at (5,-.433) {};
      \node (c3) at (6,-.433) {};
    \end{scope}
    \draw (a1) -- (a2);
    \draw (b1) -- (b2);
    \draw (c2) -- (c3) -- (c1);
  \end{tikzpicture}
  \caption{Sample motifs for $k = 2$.}
  \label{fig:motifs}
\end{figure}
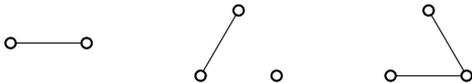

\textbf{Motif labels} To learn shared representations of bag topology, we label each bag with a motif. In this context a motif is an unlabeled subgraph induced by concepts of a bag. All permutation equivariant subgraphs map to the same motif. Given a small, fixed TD width $k$, the set of possible subgraph motifs is typically small as well (e.g., $7$ for $k = 2$). Figure~\ref{fig:motifs} displays a few example motifs, the middle of which is disconnected. The learned motif embeddings for a given AMR are denoted by $M \in \mathbb{R}^{n \times n \times d}$.

\textbf{Subgraph-induced relations} If the motif labels serve as a structural template, we are also interested in the content---a bag's relation labels. The RNN encoder over AMR paths provides a global, sequence-based view of graph structure. We augment it with bag embeddings formed as a function of component relation labels. Given a subgraph adjacency matrix $A_s \in \{0, 1\}^{k \times k}$, padded as necessary, we embed all entries besides the diagonal. $R_s \in \mathbb{R}^{k(k - 1)d}$ denotes the relation embeddings in subgraph $s$ selected from $\widetilde{R}$ (\S\ref{sec:tform-encoder}). The subgraph representation is found as
\begin{align}
  B_s = \relu\big(W_1[R_s] + \mathbf{b}_1\big).
\end{align}
Relations $(u, v) \in E$ in the same bag share a subgraph embedding: $\forall u, v \in X_i\colon S_{uv} = B_i$. Intuitively, the similarity of same-bag relation embeddings may offer valuable locality information to the attention mechanism.

\textbf{Relative bag depths} To encode position within a TD, we include a relation's bag depth relative to its source concept. As in \S\ref{sec:masked-tree} a bag's depth is the distance to the furthest leaf in its subtree. A relation assumes the depth of its enclosing bag, denoted $\tilde{a}_{ij}$ for $i, j \in [1, n]$. Source concept depth $d_i$ is the maximum depth of all relations concept $i$ is incident to. A relation's relative bag depth is
\begin{align}
  a^{(d)}_{ij} &= \abs{d_i - \tilde{a}_{ij}}.
\end{align}
This way, bag depths in row $i$ of $A^{(d)} \in \mathbb{R}^{n \times n}$ are relative to that of concept $i$. Embeddings of all such depths are stored in $D \in \mathbb{R}^{n \times n \times d}$.

The final relation embedding is a linear projection of $\widetilde{R}_{ij}$ concatenated with the preceding bag features:
\begin{align}
  \label{eq:rel-emb}
  R_{ij} &= W_2\big[\widetilde{R}_{ij}; M_{ij}; S_{ij}; D_{ij}\big] + \mathbf{b}_2.
\end{align}
The combination of structure and content bag features with their relative depths is inspired by the positional encodings of Transformer attention \cite{vaswani2017attention}. We include both $M_{ij}$ and $S_{ij}$ since bags may be similar structurally but not in terms of content (and vice versa).

\section{Experiments}
As a baseline we use the model by \citet{cai2020graph}, which encodes pairwise shortest paths between vertices to inject graph structure into self-attention. We overlay tree decomposition attention on this system to explore the effects of our structural priors. Thus, we keep the baseline's architecture the same besides applying unidirectional RNNs along the shortest paths for $\widetilde{R}$. This maintains acyclity of the tree-structured attention.

\begin{table}[t]
  \centering
  \begin{tabular}{@{}>{\quad}l c c@{}}
    \toprule
    & BLEU & \chrf{}\\
    \midrule
    \rowgr{\ldcb{}}\\
    \newcite{cai2020graph} & \textbf{27.4} & 56.4\\
    \tdmsd{} & \textbf{27.4} & \textbf{57.2}\\
    \rowgr{\ldca{}}\\
    \newcite{cai2020graph} & 29.8 & 59.4\\
    \tdmsd{} & \textbf{31.4} & \textbf{61.2}\\
    \quad\tdsd{} & 30.3 & 59.7\\
    \quad\tdmd{} & 30.4 & 60.4\\
    \quad\tdms{} & 31.0 & 61.1\\
    \bottomrule
  \end{tabular}
  \caption{Test split scores.}
  \label{tab:results}
\end{table}

Our models are run on the \ldca{} corpus of 36,521 training, 1,368 development, and 1,371 test AMR-sentence pairs. We also include results on the smaller \ldcb{} corpus with 16,833 training pairs and the same evaluation splits. We evaluate the generated sentences using BLEU \cite{papineni2002bleu} and \chrf{} \cite{popovic2017chrf++}. Models are trained on two 16GB Tesla V100 GPUs.

\subsection{Preprocessing}
To maintain consistency with the baseline, we apply the same preprocessing procedure by \citet{konstas2017neural} to concept labels and English tokens. This involves simplifying concept labels by removing sense tags (e.g., `want-01' $\rightarrow$ `want') and variable names (e.g., `y / you' $\rightarrow$ `you'). Further steps include collapsing named entity subgraphs into categorized labels and standardizing date formats. These artifacts are removed during postprocessing.

\subsection{Setup}
Among the three bag features in \S\ref{sec:bag-sg}, the relative bag depth embeddings $D$ appear to overlap the most with masked tree attention (\S\ref{sec:masked-tree}). Whereas the embeddings of motif labels $M$ and subgraph-induced relations $S$ rely only on bag internals, $D$ depends on external tree structure. In comparison with our proposed model \tdmsd{}, we ablate the three bag features in \tdsd{}, \tdmd{}, and \tdms{}.

Both models inherit the baseline's hyperparameters. Randomly initialized vertex and token embeddings are of size 300. Their encoders also apply character-level CNNs with filter size of 3 and projections of 256. All attention blocks have hidden states of size 512, feed-forward projections of size 1024, and eight attention heads. Model parameters are optimized using \textit{Adam} \cite{kingma2015adam} with default settings of $\alpha = 0.001$, $\beta_1 = 0.9$, $\beta_2 = .999$, and $\eta = 10^{-8}$.

To ensure that the relation embedding (eq.~\ref{eq:rel-emb}) size is the same across \tdms{} and \tdmsd{}, the sizes of $[S, M]$ are $[128, 128]$ in the former and those of $[S, M, D]$ are $[128, 64, 64]$ in the latter. For \tdmd{} and \tdsd{}, $[M, D]$ and $[S, D]$ have the same sizes as $[S, M]$. The halved embedding size of $M$ in \tdmsd{} relative to \tdms{} did not to hurt performance enough to counteract the benefits of $D$.

\begin{figure}[t]
  \centering
  \captionsetup[subfigure]{justification=centering}
  \subfloat[Reentrancy count.]{
    \centering
    \begin{tikzpicture}
      \centering
      \begin{axis}[name=one, width=8cm, height=5cm, legend style={font=\scriptsize}, xtick={0,1,2,3,4}, xticklabels={0,1–2,3–4,5–6,7+}, cycle list/Set1, ylabel={\small{\chrf{}}}, legend cell align={left}]
	\addplot+[only marks, mark=triangle, mark size=1.5pt] plot[error bars/.cd, y dir=both, y explicit] table [x expr={\thisrow{x}-0.25}, y=c,y error=ec,col sep=comma] {table/re_out.csv};
	\addplot+[only marks, mark=diamond, mark size=1.5pt] plot[error bars/.cd, y dir=both, y explicit] table [y=f,y error=ef,col sep=comma] {table/re_out.csv};
	\addplot+[only marks, mark=star, mark size=1.5pt] plot[error bars/.cd, y dir=both, y explicit] table [x expr={\thisrow{x}+0.25}, y=a,y error=ea,col sep=comma] {table/re_out.csv};
	\addlegendentry{\cai{}}
	\addlegendentry{\tdms{}}
	\addlegendentry{\tdmsd{}}
      \end{axis}
    \end{tikzpicture}
    \label{fig:reent-res}
  }
  \captionsetup[subfigure]{justification=centering}
  \subfloat[Diameter.]{
    \centering
    \begin{tikzpicture}
      \centering
      \begin{axis}[name=one, width=8cm, height=5cm, legend style={font=\small}, xtick={0,1,2,3,4,5,6,7}, xticklabels={0,1,2,3,4,5,6,7+}, cycle list/Set1, ylabel={\small{\chrf{}}}]
	\addplot+[only marks, mark=triangle, mark size=1.5pt] plot[error bars/.cd, y dir=both, y explicit] table [x expr={\thisrow{x}-0.25}, y=c,y error=ec,col sep=comma] {table/di_out.csv};
	\addplot+[only marks, mark=diamond, mark size=1.5pt] plot[error bars/.cd, y dir=both, y explicit] table [y=f,y error=ef,col sep=comma] {table/di_out.csv};
	\addplot+[only marks, mark=star, mark size=1.5pt] plot[error bars/.cd, y dir=both, y explicit] table [x expr={\thisrow{x}+0.25}, y=a,y error=ea,col sep=comma] {table/di_out.csv};
      \end{axis}
    \end{tikzpicture}
    \label{fig:diam-res}
  }
  \captionsetup[subfigure]{justification=centering}
  \subfloat[Treewidth.]{
    \centering
    \begin{tikzpicture}
      \centering
      \begin{axis}[name=one, width=8cm, height=5cm, legend style={font=\small}, xtick={0,1,2,3,4}, xticklabels={0,1,2,3,4}, cycle list/Set1, ylabel={\small{\chrf{}}}]
	\addplot+[only marks, mark=triangle, mark size=1.5pt] plot[error bars/.cd, y dir=both, y explicit] table [x expr={\thisrow{x}-0.25}, y=c,y error=ec,col sep=comma] {table/tw_out.csv};
	\addplot+[only marks, mark=diamond, mark size=1.5pt] plot[error bars/.cd, y dir=both, y explicit] table [y=f,y error=ef,col sep=comma] {table/tw_out.csv};
	\addplot+[only marks, mark=star, mark size=1.5pt] plot[error bars/.cd, y dir=both, y explicit] table [x expr={\thisrow{x}+0.25}, y=a,y error=ea,col sep=comma] {table/tw_out.csv};
      \end{axis}
    \end{tikzpicture}
    \label{fig:tw-res}
  }
  \caption{Test split average sentence-level \chrf{} over binned graph connectivity metrics.}
\end{figure}
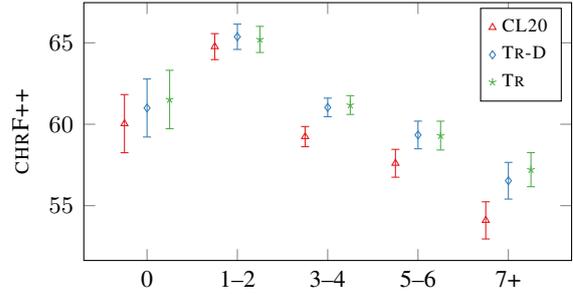
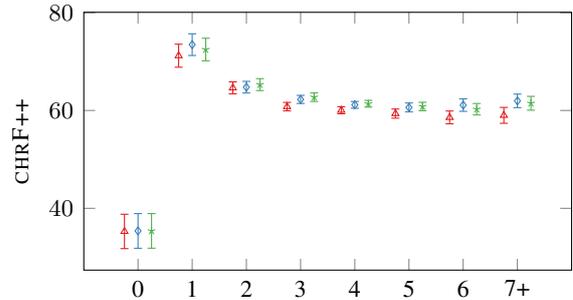
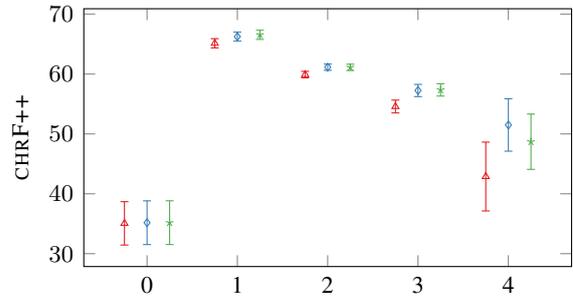

\subsection{Results}
Table~\ref{tab:results} lists performance on the two AMR corpora. On the smaller \ldcb{}, \tdmsd{} scores the same BLEU and 0.8 higher \chrf{} compared to \newcite{cai2020graph}. On \ldca{}, \tdmsd{} achieves margins of 1.6 BLEU and 1.8 \chrf{} over the baseline. \tdmsd{}'s success in both relatively low and high data settings supports its effectiveness.

Across bag features, dropping bag structure $M$ in \tdsd{} most affects generation quality with a 1.1 and 1.5 drop in BLEU and \chrf{} relative to \tdmsd{}. Next in importance is bag content, with \tdmd{} resulting in a 1.0 and 0.8 drop in the respective scores. Finally, \tdms{} only reduces BLEU and \chrf{} by 0.4 and 0.1. As relative bag depths appear to benefit the model the most, we analyze them further by comparing \tdms{} with \tdmsd{}.

\begin{table*}[t]
  \footnotesize
  \centering
  \begin{tabularx}{\textwidth}{c l X}
    \toprule
    (1) & \rf{} & the battle \textbf{horn} is about to sound , the people are already fully equipped and \textbf{ready} to go .\\
        & \cai{} & the battle is about to sound , \textbf{people are about to sound} , and people are already fully equipped to go .\\
        & \tdms{} & battle \textbf{horn} is about to sound , people are already fully equipped , and they are \textbf{ready} to go .\\
        & \tdmsd{} & the battle \textbf{horn} is about to sound , the people are already fully equipped , and the people are \textbf{ready} to go .\\
    \midrule
    (2) & \rf{} & i dont want him to not be there for his son \textbf{as he is a good father} or he at least tries .\\
	& \cai{} & i do n't want him not there with his son , or at least trying to be a good father , or at least trying to be .\\
        & \tdms{} & i do n't want him not there for his son , \textbf{because he is a good father} , or at least trying to be a good father .\\
        & \tdmsd{} & i do n't want him there without his son , \textbf{because he is a good father} , or at least trying to be a good father .\\
    \midrule
    (3) & \rf{} & take a £ 20 note on the bus , they just tell you \textbf{to get on} cos theyre lazy as hell\\
        & \cai{} & they just tell you \textbf{that they get in the bus} , take a pound 20 note on the bus , take the hell lazy .\\
        & \tdms{} & take £ 20 notes on the bus , they just tell you \textbf{to get you at the bus} , they 're hell lazy .\\
        & \tdmsd{} & ( take £ 20 notes on the bus , they will just tell you \textbf{to get into the bus} , since they are the hell lazy ) .\\
    \midrule
    (4) & \rf{} & however , the chinese team performed better today , and \textbf{therefore achieved victory} .\\
        & \cai{} & but today , the chinese team will perform better \textbf{because it 's won} .\\
        & \tdms{} & but the chinese team will be better performance today , \textbf{so it 's won} .\\
        & \tdmsd{} & but the chinese team would have better performance today , \textbf{so they won} .
\\
    \bottomrule
  \end{tabularx}
  \caption{Model-generated sentences; \rf{} is the reference and \cai{} is \citet{cai2020graph}.}
  \label{tab:sample-snts}
\end{table*}

Since tree attention includes the full subtree rooted at each bag, relative bag depths may help the attention mechanism distinguish between descendants of different depths. More generally, the bag depths also inform a query bag which of its keys are above, below, and at the same depth. We tried replacing relative tree depths with absolute ones, yet they appeared to hurt performance. Perhaps the benefits of relative tree depths on graphs are analogous to relative positions on strings.

\subsection{Analysis}
We proposed that TD structure could improve modeling ability of AMR encoders. In contrast to clique-like vertex self-attention, the acyclity of trees provide a notion of order that may correspond more closely to token sequences. To substantiate this claim, we analyze how performance varies with several graph complexity metrics: reentrancy count, diameter, and treewidth. We also study the distance between maximal-attention vertex pairs in encoder self-attention.  Furthermore, we compare our models' outputs to those of the baseline on selected sentences.

\textbf{Reentrancy} In terms of prevalence across the AMR corpus, 85.6\% of preprocessed development set AMRs featuring some form of reentrancy. We can generally view graph complexity as scaling with reentrancy count, since each reentrancy introduces a simple cycle to the undirected graph. In Figure~\ref{fig:reent-res} we plot sentence-level \chrf{} binned by increasing reentrancy counts. Both model variants \tdms{} and \tdmsd{} maintain a significant lead over the baseline, with the latter performing especially well on highly reentrant AMRs. This trend may be due to the simplified structure that tree decompositions provide. By focusing the encoder on inherent tree-like AMR structure, graph cycles may have less impact on sequence generation.

\textbf{Diameter} Another relevant metric is diameter, the maximum pairwise distance between vertices. While the baseline system only encodes shortest paths per vertex pair, our models add hierarchical structure. As a result, they may be more resilient to long-distance vertex communication costs. Figure~\ref{fig:diam-res} confirms this as the benefits of our models only grow on AMRs of higher diameters.

\textbf{Treewidth} As defined in \S\ref{sec:td-def}, treewidth quantifies the connectivity of a graph by proxy of its optimal TD's maximum bag size. As our models' encoders explicitly receive TD structure, graphs of higher treewidth should be less problematic for text generation. Figure~\ref{fig:tw-res} corroborates this view; \tdms{} and \tdmsd{} both achieve growing margins over the baseline \chrf{} as treewidth increases.

\begin{figure}
  \centering
  \begin{tikzpicture}
    \begin{axis}[colormap/GnBu,colorbar,view={0}{90},enlargelimits=false,unit vector ratio*=1 1,y dir=reverse,xlabel=\small{Head},ylabel=\small{Layer},width=7cm]
      \addplot [matrix plot*,point meta=explicit] file {table/max_attn.csv};
    \end{axis}
  \end{tikzpicture}
  \vspace{-1.5em}
  \caption{Average distance to maximal-attention key vertex per query.}
  \label{fig:attn-dist}
\end{figure}
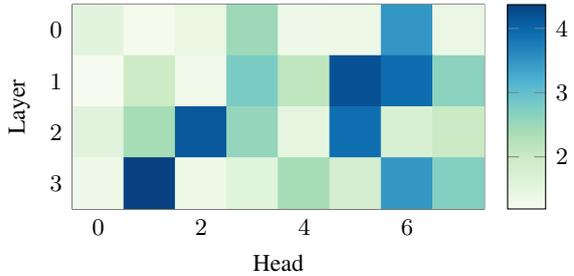

\textbf{Self-attention and graph distance} We posit that TD structure helps the graph encoder balance between local and global levels of self-attention. Following \newcite{cai2020graph}, we visualize the average distance between each query vertex and its highest weighted key on the \ldca{} development set. As seen in Figure~\ref{fig:attn-dist}, the attention heads are stratified with heads 1–2 and 5–6 preferring to attend to distant vertices. These separate roles across attention heads suggest that TDs help the encoder distinguish between close and distant vertices. Over attention layers 0--3, the mean distances are 1.78, 2.51, 2.48, and 2.38, showing that the shallowest layer favors nearby vertices.

\textbf{Manual inspection} Besides automatic evaluation we also inspect several model-generated sentences. The examples in Table~\ref{tab:sample-snts} are chosen due to their graph complexity or inclusion of traits such as causality. The AMRs of examples (2) and (3) have reentrancy counts and diameters of 4--5; those of (1) and (4) range from 2--3. All examples have a treewidth of 2, with the exception of (2) with value 3. Observe in (1) that the baseline drops the words \textit{horn} and \textit{ready} from the phrases \textit{battle horn} and \textit{ready to go}. It also suffers from repetition in producing \textit{people are about to sound}, which mistakenly echoes the preceding phrase \textit{the battle is about to sound}. These dual issues of token omission and repetition may be symptoms of underspecified graph structure, as neither \tdms{} nor \tdmsd{} make the same errors. Similarly, in (2) the baseline omits the critical phrase \textit{as he is a good father}, which is present in our models' output. Example (3) reveals more systemic errors made by the baseline; it not only fails to order the clauses correctly but uses the wrong pronouns in generating \textit{they get in the bus} when the subject is \textit{you}. Finally, (4) contains causality between the Chinese team's better performance and eventual victory. The baseline inverts the directionality between the two events by emitting \textit{because} in place of \textit{therefore}. Both our models select the word \textit{so}, which is semantically correct. That our models avoid the above issues suggests that TDs may model concept order and coverage to aid sentence generation.

\section{Conclusion}
We presented a method to bias neural graph encoders on AMR TDs. After building a parse forest of TDs, we automatically extract an `optimal' tree per AMR\@. This extraction is done by assigning local bag scores to align TD structure to that of the AMR. We proposed that TDs summarize graph structure, allowing the model to generalize over both local and global patterns in graph structure. Experiments revealed that embeddings of relative tree depth improved performance in terms of BLEU and \chrf{}. Our models also excelled on structurally complex AMRs, measured by reentrancy count, diameter, and treewidth. Qualitative analysis supports coherence of our model outputs, especially on common text generation errors such as repetition. Instead of conditioning the model one latent structure per AMR, parsing TDs within the graph encoder may improve its robustness and is left for future work.

\paragraph{Acknowledgments} This work was supported by NSF awards IIS-1813823 and CCF-1934962.

\bibliographystyle{acl_natbib}
\bibliography{acl2021}

\end{document}